# Credit Assignment in Adaptive Evolutionary Algorithms

Track: Genetic Algorithms

James M. Whitacre
School of Chemical Engineering
University of New South Wales
Sydney 2052, Australia
+612-9385 5267

z3139475@student.unsw.edu.au

Tuan Q. Pham
School of Chemical Engineering
University of New South Wales
Sydney 2052, Australia
+612-9385 5267

tuan.pham@unsw.edu.au

Ruhul A. Sarker
School of Information Technology and Electrical Engineering, University of New South Wales, ADFA Campus, Canberra 2600, Australia
+612-6268-8051

r.sarker@adfa.edu.au

**ABSTRACT**
In this paper, a new method for assigning credit to search operators is presented. Starting with the principle of optimizing search bias, search operators are selected based on an ability to create solutions that are historically linked to future generations. Using a novel framework for defining performance measurements, distributing credit for performance, and the statistical interpretation of this credit, a new adaptive method is developed and shown to outperform a variety of adaptive and non-adaptive competitors.

**Categories and Subject Descriptors**
G.1.6 **Optimization**: Stochastic programming.

## 1. INTRODUCTION
Adaptation is defined as the ability to maintain competitiveness in a changing environment. Natural adaptive systems have ingrained within them an ability to advantageously change internal components when exposed to changing external forces. This knowledge of what to change and how to change must be engineered when adaptation is considered in artificial contexts (e.g. computer experiments).

Our first step is to choose a measure (e.g. objective function value) for the effect of the changing environment. The second step is to assign the measured effect to the responsible adaptable parties. Finally, we must determine how to process this information so that the overall system can remain competitive.

In this work the adaptable components are search operator probabilities in an Evolutionary Algorithm (EA). In the following discussion, our measure of the effect of the environment is referred to as a **performance measurement**, determining the responsible adaptable parties is referred to as **credit assignment**, and the processing of information for maintaining competitiveness is referred to as **measurement interpretation**.

How we define our performance measurement and interpret the performance measurements will play important roles in the behavior of an adaptive system. A detailed discussion of these aspects and how they pertain to an adaptive EA can be found in [12] and will be briefly discussed in Section 2.2 and Section 3. However the focus of this work will be on how to assign credit to the appropriate search operators so that favorable adaptation occurs.

In this research, search operators will be adapted based on their contribution to the dynamic behavior of an EA, namely their role in the creation of offspring solutions. The performance measurement of offspring solutions will be some measure of solution fitness.

### 1.1 Credit Assignment
Since search operators act to create new offspring solutions, it seems obvious that the assessment of search operator performance will be deduced directly from the fitness of the offspring they create. This is referred to here as **Direct Credit Assignment**. However, this seemingly obvious answer to the credit assignment problem relies on an underlying premise that is challenged in this work. The challenged premise is referred to here as the **Standard Assumption** and it is outlined below within the familiar context of optimization.

#### 1.1.1 The Standard Assumption
In order to search for an optimal solution to a problem, we must make assumptions about the fitness landscape of the problem being solved. One of the most common and successfully applied assumptions is that a solution's objective function value (Fitness) can approximate a solution's usefulness in searching for more fit solutions. Following this to its logical conclusion, this implies highly fit solutions will ultimately be useful in finding the optimal solution.[1] This also implies that a solution's reproductive worth

---
[1] similar arguments also apply for use of $1^{st}$ and $2^{nd}$ derivatives of objective function

and the solution objective function value are roughly equivalent measures. For unimodal landscapes, this assumption is usually sufficient for guaranteeing an optimal solution will be found consistently and in a reasonable amount of time. However, for multimodal landscapes, this standard assumption fails to produce reliable or acceptable results.[2]

*1.1.1.1 Direct Credit Assignment*

Applying the Standard Assumption to the adaptation of search operators suggests that an operator's ability to create highly fit solutions will provide a measure for the worth of that operator. In other words, the credit for a solution is assigned directly to the operator that creates it (Direct Credit Assignment).

*1.1.2 Search Bias Assumption*

An alternative approach is to look at solving the inverse problem which is that of optimizing search bias. For clarity, we will call this optimization based on the Search Bias Assumption. Here our goal is to find solutions and search mechanisms that are most likely to reach the optimal solution in a small number of steps. Instead of assigning credit to a highly fit solution, we look to assign credit to solutions that participate in finding high fitness solutions. The underlying assumption made here is that a solution that was helpful in finding good solutions has a chance of being helpful in finding even better solutions. Furthermore, we treat this assumption as if it can be successfully applied throughout the fitness landscape all the way up to the globally optimal solution.[3] One obvious result of this approach is that a distinction is drawn between the reproductive value and the objective function value of a solution which makes this markedly distinct from optimization under the Standard Assumption.

*1.1.2.1 Credit Assignment by Search Bias*

Adhering to the Search Bias Assumption for credit assignment purposes means that credit for the performance of a current solution should be given to adaptable aspects (e.g. search operators) associated with the creation of the solution's ancestors. However, this means that little or no credit is given to the current solutions (or for that matter the most recent usage of search operators). Our inability to assign credit to "current conditions" means we have no information for adaptation purposes in the present context.

In order to obtain currently informative performance measures we must either obtain more information about the current solutions (and their future offspring) or we must make additional assumptions. Three possibilities are described below.

1. **Carry out tests**

One obvious way to assess the worth of current solutions would be to simply execute the EA in a test trial over several generations and then see which solutions were actually beneficial in future generations. Since we are dealing with a stochastic system, it is not known whether a single test trial will be indicative of the solution's expected worth and so multiple trials would be needed to ensure a solution's future worth is properly evaluated. Clearly this approach is severely limited in that we will need many function evaluations to assess a solution instead of a single evaluation as is commonly used for assessing performance. For problems where function evaluations are expensive, the costs of this approach will probably outweigh any benefits.

2. **Tests on a Simulated Model**

We might be able to get around all of the expensive function evaluations if estimates of the objective function could be used. For instance, the use of a meta-model would allow for extensive simulations which in turn might make for a more efficient approach. This alternative approach would of course be limited by the cost and accuracy of the simulations.

3. **Credit Assignment by Historical Linkage**

If we don't want to use simulations or deal with the computational burden of "multiple trials", we could simply try to work around our inability to measure the current worth of search operators. By assuming that past performance of search operators can inform us about current performance, it is possible to ignore the problem of assigning credit to current solutions while still providing measures for adaptive purposes. This is referred to as Credit Assignment by Historical Linkage and is investigated in detail in this research.

*1.1.3 Outline of Research*

This research looks at how one might assign credit to adaptable EA operators using the Search Bias assumption. In particular, we propose a method for assigning credit based on historical linkage of solutions for the purposes of adaptation of search operator probabilities.

In Section 2 a description of historical linkage is provided using a new label Event and a basic measurement concept, the Event Takeover Value. Several important credit assignment difficulties are also presented in this Section. With the credit assignment tool defined, Section 3 discusses the importance of performance interpretation in defining a competitive adaptive process. The Experimental Setup is given in Section 4 and Experimental results from several different adaptive and non-adaptive methods are presented in Section 5. Discussion and Conclusions finish off the paper in Sections 6 and 7.

## 2. CREDIT ASSIGNMENT BY HISTORICAL LINKAGE

### 2.1 Events Defined

The term Event simply refers to an instance of reproduction. That is, an Event is a specific instance of using a single search operator to generate an offspring solution. It is in essence no different than a solution except that it retains information about solutions that it is historically connected to (parents, grandparents, etc). When a new Event occurs, genealogical information is passed from parent Events to the offspring. Once a solution dies, its existence only consists of the historical information contained in its offspring.

---

[2] In EA, most selection schemes involve relaxation of the Standard Assumption. This involves treating the Standard Assumption as being true in the average sense but not strictly true (ie a probability of it being true).

[3] Like the Standard Assumption, this also has its limitations.

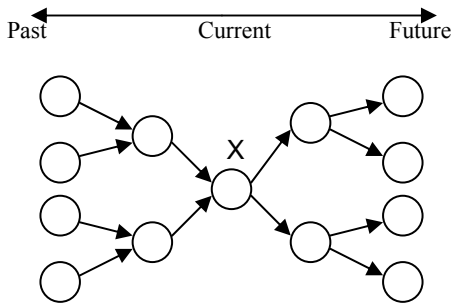

**Figure 1: Event X is shown here with the past Events which helped create X and future Events that X will help create. Events are shown as circles, arrows going into circles indicate parent Events and arrows coming out of circles indicate offspring Events.**

Given a particular Event X such as the center point in Figure 1, there will be past Events which are to some extent responsible for X's properties and there will be future Events that X will influence. In this work, the value of an Event will be propagated backward to historically linked Events.

## 2.2 Event Takeover Value

Event Takeover is a term used to define how the value of a past Event is derived from current solutions in the population. To start, the Event Takeover Value or **ETV** for an Event is simply defined as the total value of current solutions that are historically linked to the Event.

The value of each current solution will be some type of performance measurement. This is the performance measurement discussed in the Introduction and is one of the key components for enabling adaptation to occur. In this work, the performance measurement is simply a binary variable indicating whether or not a solution has survived the last selection cycle. Other measurement options are listed in [12]. Here we will define all "current" solutions to have already survived the most recent selection cycle so that each has a credit value of 1. As an example of our basic ETV measurement, if the Events on the far right of Figure 1 were members of the current population then X would be assigned an ETV of 4 (a credit of 1 for each current solution that X is connected to).

The rationale for using ETV is that the larger number of surviving Events historically connected to X, the more important the search bias associated with X. This exemplifies credit assignment by historical linkage and follows our principles laid out in the Search Bias Assumption. Starting with our basic ETV measurement, several important considerations and particular challenges for implementation are described below.

## 2.3 Information Loss from Historical Bias

To iterate from above, an Event's value is derived solely from its connection to currently surviving solutions. This means that as the current set of Events/Offspring are selectively retained or removed from the next generation, substantial information loss will take place. This is because each destroyed Event holds the historical data on linkages to past Events. If we allow the system to evolve without observing it, large amounts of information will be lost because winner's of the most recent selection processes will write the history books so to speak. This type of information loss will be referred to as **Historical Bias**. Historical Bias is a critical impediment for learning genetic linkage (gene genealogies) in the field of Population Genetics. This impediment exists in nature because we are unable to observe biological evolution as it occurs and instead can only observe the end result of evolution.

Our ability to continually monitor evolution in our computer experiments will allow us to reduce noise from Historical Bias. Evaluation of historical Events will occur after every selection process (ie every generation) and only the most favorable evaluation of a historical Event will be remembered as its ETV value. This is graphically demonstrated in Figure 2.

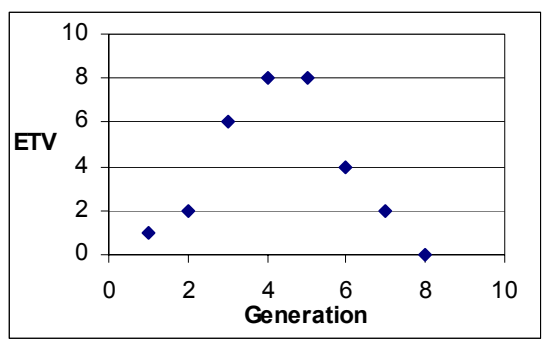

**Figure 2: This graph shows the changing ETV value of an Event from its creation to its partial takeover of the population and finally its demise. The maximum value 8 is remembered as the ETV of the Event. As an example of Historical Bias, imagine if we only observed the ETV of this Event at generation 7. Measuring at only this generation would result in this Event being highly undervalued.**

## 2.4 Genetic Dominance

In the case where an operator uses two parents, the historical information held by each new Event is double the size of the information held by the parents.[4] This doubling is computationally burdensome and makes data manipulation more difficult. It is also possible that an offspring is more similar to one of the two parents so that information might only need to be retained from a single parent. For these reasons, we looked for ways in which the historical information of only one of the parents might be retained.

Several methods for selecting the dominant parent were attempted including random selection, phenotypic similarity, and genotypic similarity between parents and offspring. In preliminary studies (results not shown), random selection resulted in mediocre EA performance as well as poor differentiation between operator probabilities. Selecting the parent that was most genetically similar (by Normalized Euclidean Distance) worked well while selecting the parent that was genetically dissimilar performed even more poorly than random selection. As a result, only historical data from the genetically most similar parent is retained when an offspring is created.[5] Phenotypic similarity appeared to provide similar results to genetic similarity however this wasn't explored in great detail.

---

[4] If one considers the arrows in Figure 1 to indicate transfer of historical information then it is easy to see how the information volume doubles with every generation.

## 2.5 Genetic Hitchhiking

In Figure 3, Events A, B and C will be given the same ETV due to the historical connection of these Events. However, it is not clear whether the success of the current population should be attributed to Events A or B since they are connected to the current population by only a single offspring (B and C, resp.). Obtaining credit based on historical linkage through a single offspring is referred to as **Genetic Hitchhiking**.

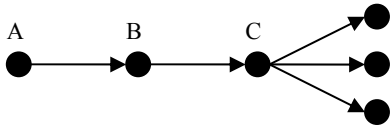

**Figure 3: Hitchhiking is shown here where credit will be passed back to Events A and B because they are historically linked to C.**

This phenomenon actually happens quite often. If an important Event occurs, it will likely spread throughout most of the population. However, all Events prior to the important Event will also spread since they are historically linked. Care must be taken then to make sure an Event has spread due to its own importance and not the importance of some later Event. To account for this, all credit is removed from an Event if it is linked to the current population by only a single future Event.

## 2.6 Credit Assignment Uncertainty

The ability to predict future states of a stochastic system is severely limited when predicting over even small time spans. Given that each discrete state change in a stochastic system is controlled by a random variable that can take several possible values, predictability decreases exponentially by the number of random variable instantiations that link two states (e.g. Events).

So far in our credit assignment description, we have assumed that the current performance of a solution can be completely and equally attributed to all historically linked events. However, since each Event is separated from the next Event by at least one random process, the confidence in our assignment of credit should quickly degrade as we look to assign credit to more distantly connected Events.[6]

---

[5] The very fact that genetic similarity was important indicates that including historical information from both parents is not necessarily the best approach. Including both parents might introduce random error due to the disproportionate importance of one parent over another in the creation of a new Offspring.

[6] The number of possible states available to the random variable will affect how quickly predictability is lost with time. Therefore it is not necessarily valid to assume that all operators

In this work, a decay schedule was used so that the proportion of credit assigned from a current solution to a historically linked Event is calculated as $\beta^x$ where $x$ is the number of search steps (or Events) separating the current solution from the historical Event. $\beta$ is referred to as the **Decay Parameter** and in this work, $\beta = 0.5$. This is a simple attempt to account for our growing uncertainty of the importance of historical linkage as that linkage becomes more distant. This schedule quickly reduces credit assigned across large distances so that a distance of $x = 6$ means that less than 0.02 of the credit of a solution will be assigned to that Event. For this reason (and to limit the amount of information stored in each Event), information passed from parent to offspring only included the last six generations ($x \leq 6$) of genealogical information.

## 2.7 Summary

The final ETV calculation is actually the result of many different calculation steps as just described. The following is a recap of these steps which provided us with our final ETV value that is used for adaptation purposes in this research.

The current population assigns credit back to historically linked events (credit assignment by historical linkage). Credit is only assigned back to dominant parent Events as defined by genotypic similarity (ie *Genetic Dominance*). The credit itself becomes exponentially smaller as it passes back to older historical events (ie *Credit Assignment Uncertainty*). After all credit has been passed back, any events with only a single link tying it to the current population are removed from consideration (ie *Genetic Hitchhiking*). Finally, the dynamic value of Event credit is continually observed and only the most favorable assessment of an Event is retained as the final measurement of the Event's value (ie *Historical Bias*). The application of these steps is highlighted in the Pseudo Code below.

**Pseudo Code for Credit Assignment by Historical Linkage (for running a single generation of the adaptive EA)**

- Reproduction (ie Event Creation) of all offspring
  - A unique Event ID is assigned to each offspring.
  - Offspring takes information on historical linkage (Event ID info) from genetically dominant parent (and only information from the last six historical Events is kept).
- Selection (Select next generation from Parents + Offspring)
- Credit Assignment Method
  - Assign Credit from each current solution back to historically linked Events.
    - Credit exponentially decays based on historical separation of Event from current solution.
    - Each time credit is added to an Event, remember the offspring Event that provided a link to the current solution.
  - If all the credit assigned to an Event came via a single offspring Event then Event credit = 0.
  - Check each Event for its existence in the Event Archive
    - If Event is new then add to Archive (Add Event ID# and value). Otherwise, determine if the new Event value is larger than the Archived value and retain the larger value.

---

will lose predictability at equal rates. For simplicity, this was neglected in this work.

## 2.8 Similar Research

Although all aspects of ETV were derived independently, the use of historical linkage for assigning credit and credit assignment uncertainty were proposed (using different labels) by Davis [2] and also used by [1], [7]. In the next section, the process by which ETV is used for adaptation is discussed.

## 3. STATISTICAL INTERPRETATION OF PERFORMANCE

Before we use our ETV measurement for adaptive purposes, it is helpful to consider how we want to interpret the ETV measurements. In this work, statistical methods are used to determine the extent that measurements are outliers which in turn is used as a final measure for adaptation. The purpose of focusing on outliers is to assess an operator's potential to create solutions of extraordinary reproductive value instead of looking at the average reproductive value.

Figure 4 is provided below to illustrate how the use of outliers can influence the interpretation of an operator's performance. Probability distribution functions for data from two hypothetical search operators is given with one intending to represent an explorative operator (▲) and one an exploitive operator (●). Explorative search is expected to overwhelmingly produce Events with low ETV values yet have the potential to occasionally create Events that eventually lead to highly competitive offspring (ie high ETV value) in new portions of the solution space. Exploitive search on the other hand is expected to create Events that consistently are able to spread to some limited degree but rarely lead to "groundbreaking" Events.

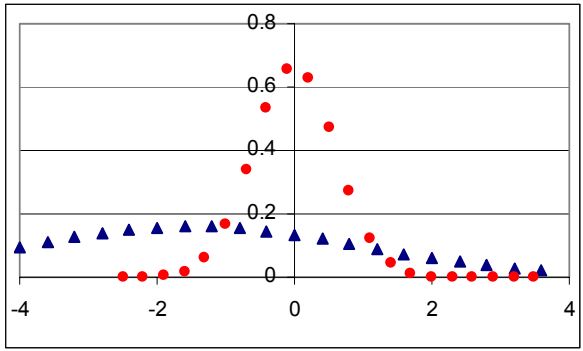

**Figure 4**: **Hypothetical probability distribution functions for offspring (ie Event) ETV values from an explorative search operator (▲) and an exploitative search operator (●).**

If adaptation is based on mean ETV measurements, we expect an adaptive method to prefer searches characterized by the (●) distribution in the Figure above. However, an operator's potential to lead to high fitness offspring in future generations (Optimization by Search Bias) is unlikely to be indicated by its average ETV. Instead we should promote those search operators that are capable of occasionally making greater leaps forward such as the (▲) distribution. We believe that the selection of outliers for adaptation can at the very least allow search mechanisms characterized by the (▲) distribution to compete with search mechanisms characterized by the (●) distribution.

In this work, the interpretation of performance by statistical outlier detection is labeled as I:3 (see Table 3) while using the average of performance measures is labeled as I:1.

For more details on statistical outlier determination and the associated calculation steps, see [12]. In the next section, the primary attributes of the EA design used for experimentation purposes is described. Other aspects of the experimental conditions are also discussed.

### 3.1 Adaptation Cycle Implementation

Each adaptive method involves the adaptation of all search operator probabilities with probability values updated every 20 generations. Also, a minimum probability of 0.02 is imposed to ensure a small number of measurements continue to be taken for the worst operators. Probability values are updated so that the new value is 50% from the previous value and 50% is from the most current adaptive cycle. All probability values are initialized at equal values unless otherwise stated.

**Pseudo Code for Adaptation**

- Every 20 generations, evaluate stored values (e.g. ETV) in Archive based on statistical outlier detection (I:3) or average value (I:1) for final performance measure of search operators.
- Normalize search operator performance measures so that they sum to one. Search operator probability is modified so that half of it is taken from previous probability value and half from the normalized performance measure.
- If Operator probability < 0.02 then probability = 0.02
- Purge Archive of ETV measurements

## 4. EXPERIMENTAL SETUP

### 4.1 Search Operators

Table 1: The 10 search operators used are listed here including name, reference for description, and parameter settings if different from reference. Operators 7 and 8 are defined below instead of referenced.

| ID | Name | Parameter Settings | Ref. |
|----|------|-------------------|------|
| 1 | Wright's Heuristic Crossover | $r = 0.5$[7] | [4] |
| 2 | Simple Crossover | — | [4] |
| 3 | Extended Line Crossover | $\alpha=0.3$[8] | [4] |
| 4 | Uniform Crossover | — | [4] |
| 5 | BLX-$\alpha$ | $\alpha=0.2$ | [4] |
| 6 | Differential Operator | | [11] |
| 7 | Swap | — | — |
| 8 | Raise | $A = 0.01$ | — |
| 9 | Creep | $A = 0.001$[9] | [9] |
| 10 | Single Point Random Mutation | — | [4] |

---

[7] $r$ is set to a static value instead of being a random variable as in the original description

[8] $\alpha$ is set to a static value instead of being a random variable as in the original description

[9] Only a single gene is randomly selected instead of performing operation on all genes.

Ten search operators were used in all adaptive EA designs as listed in Table 1. Two of the operators were original creations and are described below.

**Swap**: Select the most dissimilar gene between two parents. Transfer all genes from the better fit parent to the offspring except for the previously selected gene which is taken from the less fit parent.

**Raise**: This is similar to Creep except all genes are shifted instead of a single gene. The size of the shift is proportional to the size of each gene's range with $A = 0.01$.

## 4.2 Core EA Design
A real coded EA was used with population size of 30, population solution uniqueness enforced, and binary tournament selection with replacement for both mating (parent) and culling (parent + offspring). Reproduction consisted of the probabilistic use of a single search operator with search operator probabilities normalized to one. Populations were randomly initialized and the stopping criteria was set as a maximum number of generations. All test functions were transformed (if necessary) to be maximization problems with optima at 0. The global optima was assumed to be reached for objective function values > -1E-15.

## 4.3 Test Problems
Experiments were conducted on 10 test problems which are listed in Table 2. All of the test problems used are defined over continuous variable domains with simple bounded constraints on the variables (ie convex search space). The test problems are also characterized as being static with a single objective function.

**Table 2: Test Problems are listed with identification number, common name, and number of dimensions (variables). More information on each test problem can also be found in the stated reference.**

| ID | Name | Variables | Reference |
|---|---|---|---|
| F1 | Shekel 's Foxholes | 2 | [3] |
| F2 | Rastrigin | 20 | [8] |
| F3 | Schwefel | 10 | [8] |
| F4 | Griewank | 10 | [8] |
| F5 | Bohachevsky | 2 | [5] |
| F6 | Watson's | 5 | [5] |
| F7 | Colville's | 4 | [5] |
| F8 | System of linear equations | 10 | [5] |
| F9 | Ackley's | 25 | [5] |
| F10 | Neumaier's #2 | 4 | [6] |

## 4.4 Diversity Control
Single point Mutation (Operator 10) was used for maintaining population diversity. The probability of using operator 10 was set using a deterministic approach proposed by Pham [10] where the probability is exponentially related to the distance $d$ between parents $A$ and $B$. $\delta$ is a parameter that can be tuned and in this work, $\delta = 0.001$ for all test problems.

$$P_{Mut} = P_{Mut}^o + 0.5^{\frac{d}{\delta}} \qquad (1)$$

$$d^2 = \sum_{i=1}^{N_{var}} \left( \frac{x_i^A - x_i^B}{x_{i,max} - x_{i,min}} \right)^2 \qquad (2)$$

In Equation 2, a solution is represented as a vector of search variables $x$ with the $i^{th}$ variable having upper and lower bounds, $x_{i,max}$ and $x_{i,min}$.

## 4.5 Suite of Algorithms Tested
**Table 3: Names and adaptive characteristics of EA designs.**

| Name | Measurement Interpretation Section 3 | Diversity Control Section 4.4 | Credit Assignment (direct/ETV) |
|---|---|---|---|
| EA1 | I:1 | No | direct |
| EA2 | I:3 | No | direct |
| EA3 | I:1 | Yes | direct |
| EA4 | I:3 | Yes | direct |
| EA5 | I:1 | No | ETV |
| EA6 | I:3 | No | ETV |
| EA7 | I:1 | Yes | ETV |
| EA8 | I:3 | Yes | ETV |
| SGA | N/A | Yes | N/A |

The nine EA designs tested in this work are listed in Table 3. In addition to the adaptive EA designs considered (EA1 through EA8) we also included a simple GA (SGA) which uses only two operators: Operator 4 with probability of 0.98 and Operator 10 with probability set by Diversity Control ($P^0 = 0.02$).

Also, an important point must be mentioned regarding the use of I:3. When attempting to apply an outlier detection method such as I:3, it is important to recognize that not all measurements are capable of producing outliers. For instance, the binary performance measurement used throughout this work will not produce outlier values when coupled with direct credit assignment. As a result, adaptation won't occur for each of the adaptive methods using I:3 with direct credit assignment. This applies to EA designs EA2 and EA4.

## 5. EXPERIMENTAL RESULTS
### 5.1 Assessing EA performance
The performance of a single run of an EA is typically given as the best solution fitness found. Since an EA is a stochastic search process, we must conduct several runs and then extract useful information from the resulting sample of performance data. It is common practice to compare different EA designs based on the sample's average performance and also based on the overall best solution found in a sample. In our results, we decided instead to use a statistical test which measures our confidence that a given EA design is superior to its competitors.

With each EA design being executed 10 times on a test problem, the data set (of EA performances) can be treated as being sampled from a distribution which in turn can be compared with other EA design data sets. Most of the data sets did not fit standard parametric (e.g. $z$ test) distributions as indicated by probability plots and so the Mann-Whitney U Test[10] was used, which doesn't assume a particular distribution.

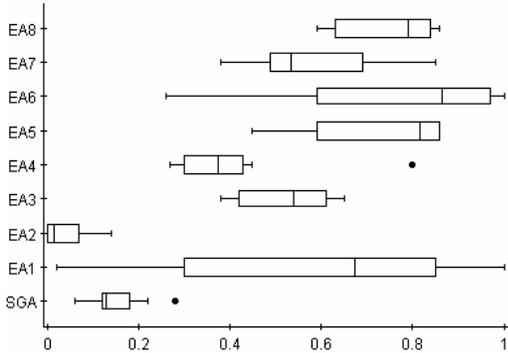

**Figure 5: Boxplot of Mean performance measures for each EA design over all 10 test functions.**

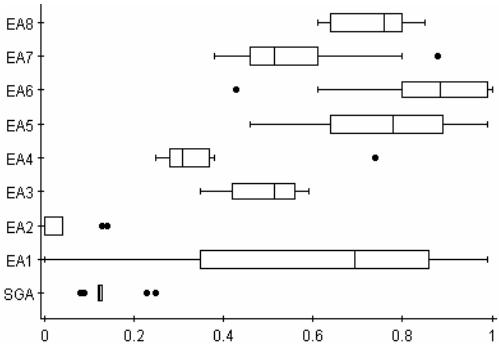

**Figure 6: Boxplot of Final performance measures for each EA design over all 10 test functions.**

Using a one-sided Mann-Whitney U Test, confidence levels (1 - $p$ value) are calculated to indicate whether one sample is greater than another sample. The average confidence level is then used as our measure of EA performance and indicates our expectation that a particular EA design is better than the other competing designs.

*5.1.1 Stopping Criteria*
It is also common for experimental results to state performance for a single stopping criteria. If we want to make general statements about the usefulness of an EA then the selection of a single stopping point will clearly introduce bias into the experimental results as well as the associated conclusions. In an attempt to minimize this bias, stopping points were considered at every 100 generations with a final stopping point at 2000 generations. From the 20 stopping points, two important pieces of information have been extracted and are presented in this section.

First, the average performance over all stopping points (**Mean**) for each EA design is presented in Figure 5. This allows us to see if the EA performed well consistently throughout execution. Our second measure is simply the performance measure at the final stopping point (**Final**) and is presented in Figure 6. This provides us with some indication of long-term performance.

---

[10] Other non-parametric tests are available and would have also been valid.

## 5.2 Factorial Analysis of Adaptive Components

A factorial analysis was conducted to determine the main effects and first level interactions between several design components associated with the adaptive systems studied. The parameters Credit Assignment (I:1/I:3), Diversity Control (inactive/active), and Measurement Interpretation (Direct Credit Assignment/Historical Linkage by ETV) were coded using the standard convention (-1/1). The output measures used for this analysis were the Mean (Table 4) and Final (Table 5) performance measures. The experimental design was blocked for test problem.

**Table 4: Factorial analysis for Mean performance**

| Factor  | Effect | t Stat | P-value  |
|---------|--------|--------|----------|
| I:3     | -5.81  | -2.67  | 9.41E-03 |
| Div     | 1.99   | 0.91   | 3.65E-01 |
| ETV     | 15.86  | 7.28   | 3.18E-10 |
| I:3*Div | 6.94   | 3.19   | 2.13E-03 |
| I:3*ETV | 10.89  | 5.00   | 3.87E-06 |
| Div*ETV | -5.65  | -2.59  | 1.15E-02 |

**Table 5: Factorial analysis for Final performance**

| Factor  | Effect | t Stat    | P-value  |
|---------|--------|-----------|----------|
| I:3     | -5.52  | -2.66656  | 0.009431 |
| Div     | -0.79  | -0.38373  | 0.702296 |
| ETV     | 17.92  | 8.652576  | 8.36E-13 |
| I:3*Div | 6.04   | 2.915375  | 0.004715 |
| I:3*ETV | 11.66  | 5.62747   | 3.19E-07 |
| Div*ETV | -6.57  | -3.17363  | 0.002204 |

## 6. DISCUSSION
### 6.1 ETV Measurements
No single EA design was able to outperform all others for every test function and stopping criteria considered however there were some obvious trends in the experimental results provided in the last Section. Clearly the most important effect on performance was the use of credit assignment by Historical Linkage which proved to have a strong advantage over direct credit assignment (Factor "ETV" in Table 4 and Table 5). We can also see that the performance improvements from using Historical Linkage were strong and fairly consistent, as indicated by the box plots for EA5, EA6, EA7, and EA8 in Figure 5 and Figure 6.

### 6.2 ETV + Outlier Detection
Almost as important however is the combined use of Historical Linkage credit assignment with the statistical selection of outlier ETV values (Factor "I:3*ETV" in Table 4 and Table 5).

The ETV measurement was designed so that it could "see" over several search steps, placing value on a solution based on the ability of its lineage to survive. When dealing with any EA with fixed, finite population size, it is expected that most lineages perish to make room for a select few that flourish. In order to select operators that tend to produce those infrequent flourishing lineages, it is necessary to pay special attention to those infrequent occurrences instead of paying attention to averages. The statistical outlier detection of I:3 fulfills this need by interpreting measurement worth based on the degree in which it is an outlier. The combination of I:3 with ETV performs very nicely as observed in the results of EA6 and EA8 (Figure 5 and Figure 6). Without the I:3 statistical interpretation of ETV, the ETV credit assignment approach has significantly less impressive performance as seen in EA5 and EA7 (Figure 5 and Figure 6).

### 6.3 Outlier Detection
It is interesting to note that I:3 was found on average to have a negative impact on performance. As previously mentioned, I:3 essentially halts adaptation when used with direct credit assignment and a low resolution performance measure. This occurs for both EA2 and EA4 and the poor performance of these two EA designs is likely to be the primary reason why a negative Effect is observed for I:3. Other research using I:3 with higher resolution performance measurements has indicated that in situations where measurement outliers will occur, that I:3 actually improves performance of the EA [12].

### 6.4 Limits to the utility of Search Bias
Although adaptive methods using ETV were often the best performing EA designs, it should be noted that very little credit is assigned from solutions to Events separated by more than a few search steps (a result of the decay parameter described in Section 2.6). Since we are dealing with stochastic systems, we know that predictability is lost after a short number of steps and so the degree to which the Search Bias Assumption can be exploited is also expected to be limited as we tried to account for by using the decay parameter.

Our results suggest that use of the Search Bias Assumption (via Credit Assignment by Historical Linkage) can provide useful information for adaptation purposes however its limits need to be recognized. Exploring the nature of these limitations in more detail seems to be a worthwhile goal both in creating more competitive EA and in understanding evolutionary processes.

### 6.5 ETV Implementation
Under the somewhat limited range of parameter combinations tested (results not shown), our preliminary testing suggested that Genetic Dominance and Historical Bias were the two most important factors in improving performance over the raw ETV measurement. Hitchhiking and Credit Assignment Uncertainty were not as important to EA performance and yet the latter was hypothesized to play a very important role. The reason for their marginal influence is not immediately obvious and will be looked at in more detail in future research.

## 7. CONCLUSIONS
The process by which we assign reproductive value to solutions (credit assignment) can play an important role in adaptive processes. A philosophical framework for optimization based on search bias was proposed and shown to be effective in adaptation of search operator probabilities when credit for a solution's fitness was assigned through historical linkage. When combined with historical linkage, the statistical selection of outliers for interpreting performance was also shown to play an important positive role in adaptation.